\documentclass[conference]{IEEEtran}

\makeatletter
\def\ps@headings{%
\def\@oddhead{\mbox{}\scriptsize\rightmark \hfil \thepage}%
\def\@evenhead{\scriptsize\thepage \hfil \leftmark\mbox{}}%
\def\@oddfoot{}%
\def\@evenfoot{}}
\makeatother
\usepackage{amssymb}
\usepackage{amsmath}
\usepackage{amsthm}
\usepackage{amssymb}
\usepackage{graphicx,tikz}
\usetikzlibrary{arrows,automata}

\usepackage{cite}
\usepackage{epsfig,psfrag,subfigure}
\usepackage{amsmath,amssymb}
\usepackage{array}

\begin{document}

\title{BPRS: Belief Propagation Based Iterative Recommender System}

\author{ Erman Ayday, Arash Einolghozati, Faramarz Fekri\\
School of Electrical and Computer Engineering \\ Georgia Institute of Technology, Atlanta, GA 30332\\
\texttt{Email:}\{eaydayi, einolghozati, fekri\}@ece.gatech.edu}

\maketitle
\thispagestyle{empty}
\pagestyle{empty}

\begin{abstract}
In this paper we introduce the first application of the Belief Propagation (BP) algorithm in the design of recommender systems. We formulate the recommendation problem as an inference problem and aim to compute the marginal probability distributions of the variables which represent the ratings to be predicted. However, computing these marginal probability functions is computationally prohibitive for large-scale systems. Therefore, we utilize the BP algorithm to efficiently compute these functions.  Recommendations for each active user are then iteratively computed by probabilistic message passing. As opposed to the previous recommender algorithms, BPRS does not require solving the recommendation problem for all the users if it wishes to update the recommendations for only a single active. Further, BPRS computes the recommendations for each user with linear complexity and without requiring a training period. Via computer simulations (using the $100$K MovieLens dataset), we verify that BPRS iteratively reduces the error in the predicted ratings of the users until it converges. Finally, we confirm that BPRS is comparable to the state of art methods such as Correlation-based neighborhood model (CorNgbr) and Singular Value Decomposition (SVD) in terms of rating and precision accuracy. Therefore, we believe that the BP-based recommendation algorithm is a new promising approach which offers a significant advantage on scalability while providing competitive accuracy for the recommender systems.
\end{abstract}
\vspace{-.1in}

\section{Introduction}
\label{sec:intro}

Today, the quantity of available information grows rapidly, overwhelming consumers to discover useful information and filter out the irrelevant items. Thus, the user is confronted with a big challenge of finding the most relevant information or item in the short amount of time.  Recommender systems are aimed at addressing this overload problem, suggesting to the users those items that meet their interests and preferences.  More generally, recommender systems can learn about user preferences and profile over time, based on data mining algorithms, and automatically suggest products (from a large space of possible options) that fit the users' needs. Hence, it is foreseeable that the social web is going to be driven by these recommender systems.

However, there are certain challenges to design scalable, accurate and dependable recommender systems. The available data for the recommender systems is incomplete, uncertain, inconsistent and/or intentionally-contaminated. Further, since new data (ratings) becomes available continuously, recommendations need to be updated in frequent intervals causing computational limitations for large-scale systems. Latent factor models (such as Matrix Factorization) have proven to be the most accurate method in the Root Mean Square Error (RMSE) sense. However, most existing and highly popular Matrix Factorization-based recommender algorithms are shown to be prone to malicious behavior~\cite{Cheng_recsys09} and they have scalability issues. In other words, they fall short of incorporating the attack profiles and the extra noise generated by the malicious users. Further, each new update (using the most recent data or ratings) for a particular active user requires to solve the entire problem for every user in the system. Hence, new research needed to focus on algorithms which meet these challenges and provide scalable, accurate and dependable recommender systems.

In this paper we introduce the first application of Belief Propagation (BP), an iterative probabilistic algorithm, to solve the recommendation problem. We have applied BP to trust and reputation systems in our previous work~\cite{Ayday_ISIT2011,ayday_TDSC_12}. In such systems, BP is used to solve the inference problem for finding the global reputation of service providers in a network based on the previous ratings of the users. The main difference between trust and reputation systems and recommender systems is that in the former one the inference problem has to be solved globally but in the latter one, the inferences are local and specific for each user. In~\cite{ayday_TMC_12} and~\cite{ayday_SECON_12}, we have studeied the reputation system for Delay tolerant networks (DTN) and P2P networks respectively.

The key observation we make is that recommender systems deal with complicated global functions of many variables (e.g., users and items). By using a factor graph, we can obtain a qualitative representation of how the users and items are related on a graphical structure. Therefore, we propose to model the recommender system on a factor graph using which our goal is to compute the marginal probability distribution functions of the variables representing the ratings to be predicted for the users. However, we observe that computing the marginal probability functions is computationally prohibitive for large-scale recommender systems. Therefore, we utilize the BP algorithm to efficiently compute these marginal probability distributions. The key role of the BP algorithm is that we can use it to compute the marginal distributions in a complexity that grows linearly with the number of nodes (i.e, users/items).

Hereafter, we refer to our scheme as the ``Belief Propagation Based Iterative Recommender System'' (BPRS). BPRS has several prominent features. First, it does not require to solve the problem for all users if it wishes to update the predictions for only a single active user and it does not require a training period to utilize the most recent data (ratings). Second, its complexity remains linear per single user, making it very attractive for large-scale systems. Therefore, it can update the recommendations for each active (online) user instantaneously using the most recent data (ratings). Further, we show that BPRS provides comparable usage prediction and rating prediction accuracy to other popular methods such as the Correlation-based neighborhood model (CorNgbr) and Singular Value Decomposition (SVD). Therefore, we are very optimistic that this work promises a new direction for the recommender systems which will be scalable, accurate, and resilient to attacks.

The rest of this paper is organized as follows. In the rest of this section, we summarize the related work. In Section~\ref{sec:BPRS}, we describe the proposed BPRS in detail. Next, in Section~\ref{sec:evaluation}, we evaluate BPRS via computer simulations using the MovieLens dataset. Finally, Section~\ref{sec:conclusion} concludes the paper.

\subsection{Related Work}\label{sec:related_work}
Recommender systems~\cite{Resnick_recsys} can be classified into two main categories: i) content-based filtering~\cite{Balabanovic97fab:content-based} in which the system uses behavioral data about a user to recommend items similar to those previously consumed by the user, and ii) collaborative filtering~\cite{Resnick_grouplens} in which the system compares one user's behavior against the other users' behaviors and identifies items which were preferred by similar users. Collaborative filtering algorithms fall further into two general classes: memory-based~\cite{Herlocker02} and model-based algorithms~\cite{Hofmann99latentclass,Koren_2008}. Model-based algorithms include methods exploiting Singular Value Decomposition (SVD), Principal Component Analysis (PCA) and Maximum Margin Matrix Factorization (MMMF) techniques~\cite{Sarwar00applicationof,MMMF}.

The application of Bayesian networks and message passing algorithms for recommender systems is also studied in the past~\cite{Stern_matchbox,Campos_2008}. In~\cite{Stern_matchbox}, the message passing technique is used to determine the latent factors of the users and items (as an alternative to SVD). In~\cite{Campos_2008}, because of the fuzziness associated with the ambiguity in the description of the ratings, a (non-iterative) inference is proposed among the users to remove this ambiguity. The key difference between our approach and the other message passing-based methods is that, we describe the recommendation problem as computing marginal likelihood distributions from complicated global functions of many variables and to use Belief Propagation (BP) to find them. This is inspired by successful applications of BP algorithms in various fields such as decoding of error correcting codes~\cite{KFL-IT-2001}, Artificial Intelligence~\cite{Pear-1988}, and reputation systems~\cite{Ayday_ISIT2011}.

\section{Belief Propagation for \\Recommender Systems}\label{sec:BPRS}

Belief Propagation (BP)~\cite{Pear-1988,KFL-IT-2001} is a message passing algorithm for performing interface on graphical models (e.g., factor graphs, Bayesian networks, Markov random fields). It has demonstrated empirical success in numerous applications including LDPC codes, turbo codes, free energy approximation, and satisfiability. BP is a method for computing marginal distributions of the unobserved nodes conditioned on the observed ones.

 Our objective is to formulate the recommendation problem as making statistical inference about the ratings of users for unseen items based on observations. That is, given the past data evidence, what would be the likelihood (probability) that the rating takes a particular value? Here, the probability is the degree of belief to which the prediction of the rating is supported by the available evidence. This requires finding the marginal probability distributions of the variables representing the ratings of the items to be predicted conditioned on some observed preferences.

We assume two different sets in the system: i) the set of users $\mathbb{U}$ and ii) the set of items (products) $\mathbb{I}$. Users provide feedbacks, in the form of ratings, about the items for which they have an opinion. The main goal is to provide accurate recommendations for every user by predicting the ratings of the user for the items that he/she has not rated before (unseen item). Here, we consider an arbitrary user $z$ (referred as the active user) and compute the prediction of ratings for user $z$ for unseen items.
We assume $u$ users and $s$ items in the system (i.e., $|\mathbb{U}|=u$ and $|\mathbb{I}|=s$). Let $\mathbb{G}_z=\{G_{zj}:j\in\mathbb{I}\}$ be the collection of variables representing the ratings of the items to be predicted for the active user $z$. Note that a subset of these variables are already known as the corresponding items were rated by user $z$. Hence, they do not require any prediction. Let also $\mathbb{R}_z=\{R_{zi}:i\in\mathbb{U}\}$ be the confidence of the system on the users for their ratings' reliability, given the active user is $z$. Further, we let $T_{ij}$ represent the rating provided previously by user $i$ about the item $j$. We denote $\mathbb{T}$ as the $s\times{u}$ item-user matrix that stores these ratings, and $\mathbb{T}_i$ as the set of ratings provided by the user $i$. We note that some rating entries could be missing (attributed to unseen items). To be consistent with the most of existing recommender systems, we assume that the rating values are integers from the set $\Upsilon=\{1,2,3,4,5\}$.

The recommendation problem can be viewed as finding the marginal probability distributions of each variable in $\mathbb{G}_z$, given the observed data (i.e., existing ratings and the confidence of the system for the user's ratings). There are $s$ marginal probability functions, $p(G_{zj}|\mathbb{T},\mathbb{R}_z)$, each of which is associated with a variable $G_{zj}$; the predicted rating of item $j$ for user $z$. We formulate the problem by considering the global function $p(\mathbb{G}_z|\mathbb{T},\mathbb{R}_z)$, which is the joint probability distribution function of the variables in $\mathbb{G}_z$ given the rating matrix and the confidence of the system for the user's ratings. Then, clearly,
each marginal probability function $p(G_{zj}|\mathbb{T},\mathbb{R}_z)$ may be obtained as follows: \vspace{7pt}
\begin{equation}
p(G_{zj}|\mathbb{T},\mathbb{R}_z) = \sum\limits_{\mathbb{G}_z\backslash\{G_{zj}\}}p(\mathbb{G}_z|\mathbb{T},\mathbb{R}_z),
\label{eqn:joint_dist}
\vspace{7pt}
\end{equation}
where the notation $\mathbb{G}_z\backslash\{G_{zj}\}$ implies all variables in $\mathbb{G}_z$ except $G_{zj}$.

Unfortunately, the number of terms in (\ref{eqn:joint_dist}) grows exponentially with the number of variables, making the direct computation infeasible for large-scale systems. However, we propose to factorize (\ref{eqn:joint_dist}) to local functions $f_i$ using a factor graph and utilize the BP algorithm to calculate the marginal probability distributions in linear complexity. A factor graph is a bipartite graph containing two sets of nodes (corresponding to variables and factors) and edges incident between two sets. Following~\cite{KFL-IT-2001}, we form a factor graph by setting a variable node for each variable $G_{zj}$, a factor node for each function $f_i$, and an edge connecting variable node $j$ to the factor node $i$ if and only if $G_{zj}$ is an argument of $f_i$.

\begin{figure}[ht]
   \centering
   \epsfig{width=.7\linewidth,file=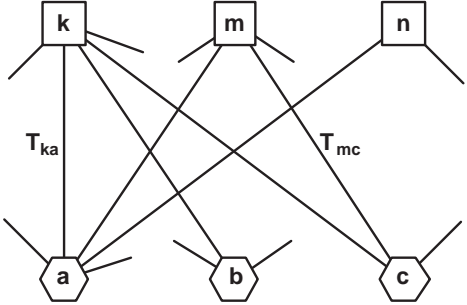}
   \caption{\footnotesize{Graphical representation of the scheme from user $z$'s point of view.}}
   \label{fig:recommender}
   \vspace{0pt}
\end{figure}

We arrange the collection of the users and items together with the ratings provided by the users as a factor graph $g(\mathbb{U},\mathbb{I})$. Then, since we consider the particular active user $z$, the factor graph is reduced to $g(\hat{\mathbb{U}},\mathbb{I})$ (as in Fig.~\ref{fig:recommender}) by only keeping the users that are connected to $z$ via a path of length at most two in $g(\mathbb{U},\mathbb{I})$ (i.e., the users who rated at least one item that is also rated by $z$) and removing all the other user nodes from the graph together with their edges. In this representation, each user corresponds to a factor node in the graph, shown as a square and each item is represented by a variable node shown as a hexagon. Further, each rating is represented by an edge from the factor node to the variable node. Hence, if a user $i$ ($i\in\hat{\mathbb{U}}$) has a rating about item $j$ ($j\in\mathbb{S}$), we place an edge with value $T_{ij}$ from the factor node $i$ to the variable node representing item $j$. Eventually, the $g(\hat{\mathbb{U}},\mathbb{I})$ graph has $|\hat{\mathbb{U}}|=\hat{u}$ users and $|\mathbb{I}|=s$ items.

Next, we suppose that the global function $p(\mathbb{G}_z|\mathbb{T},\mathbb{R}_z)$ factors into products of several local functions, each having a subset of variables from $\mathbb{G}_z$ as arguments as follows:\vspace{7pt}
\begin{equation}
p(\mathbb{G}_z|\mathbb{T},\mathbb{R}_z)=\frac{1}{Z}\prod\limits_{i\in\hat{\mathbb{U}}}f_i(\mathcal{G}_{zi},\mathbb{T}_i,R_{zi}),
\label{eqn:factorization1}
\vspace{7pt}
\end{equation}
where $Z$ is the normalization constant and $\mathcal{G}_{zi}$ is a subset of $\mathbb{G}_z$. Hence, in the graph representation of Fig.~\ref{fig:recommender}, each factor node is associated with a local function and each local function $f_i$ represents the probability distributions of its arguments given the confidence of the system for the associated user and the existing ratings of the associated user.

We now describe the message exchange between a user $k$ and an item $a$ (in Fig.~\ref{fig:recommender}) provided that the active user is $z$ in BPRS. We clarify that all the messages are formed by the algorithm that is ran in the central authority. We represent the set of neighbors of the variable node $a$ and the factor nodes $k$ and $z$ (in $g(\hat{\mathbb{U}},\mathbb{I})$) as $\mathbf{N_a}$, $\mathbf{N_k}$, and $\mathbf{N_z}$, respectively (neighbors of an item are the set of users who rated the item while neighbors of a user are the items which it rated). Further, let $\Xi = \mathbf{N_a} \backslash \{k\}$ and $\Delta = \mathbf{N_k} \backslash \{a\}$. Let $G_{zj}^{(\nu)}$ and $R_{zi}^{(\nu)}$ be the value of variable $G_{zj}$ and system's confidence on user $i$ at the iteration $\nu$ of the algorithm, respectively. The message $\lambda^{(\nu)}_{k\rightarrow{a}}(G_{za}^{(\nu)})$ (from factor node $k$ to the variable node $a$) denotes the relative probabilities that $G_{za}^{(\nu)}=\ell$ ($\ell\in\Upsilon$) at the $\nu^{th}$ iteration, given $T_{ka}$ and $R_{zk}^{(\nu-1)}$. On the other hand, $\mu^{(\nu)}_{a\rightarrow{k}}(G_{za}^{(\nu)})$ (from variable node $a$ to the factor node $k$) denotes the probability that $G_{za}^{(\nu)}=\ell$ ($\ell\in\Upsilon$) at the $\nu^{th}$ iteration.

\begin{figure}[t]
   \centering
   \small
   \psfrag{?}{$\lambda$}
   \epsfig{width=\columnwidth,file=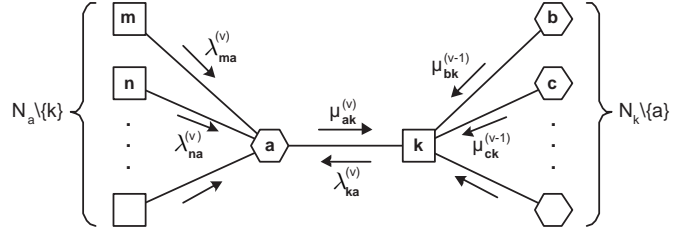}
   \caption{Message exchange between the factor node $k$ and variable node $a$.}
   \label{fig:message_xchange}
   \vspace{-5pt}
\end{figure}

The message from the factor node $k$ to the variable node $a$ at the $\nu^{th}$ iteration is formed using the principles of the BP as
\begin{equation}
    \lambda^{(\nu)}_{k\rightarrow{a}}(G_{za}^{(\nu)}) = \hspace{-15pt} \sum\limits_{\mathcal{G}_{zk}^{(\nu)}\backslash\{G_{za}^{(\nu)}\}}\hspace{-10pt}f_k(\mathcal{G}_{zk}^{(\nu)},\mathbb{T}_k,R_{zk}^{(\nu-1)})\prod\limits_{x\in\Delta}\mu^{(\nu-1)}_{x\rightarrow{k}}(G_{zx}^{(\nu)}),
    \label{eqn:lambda_zero}
    \vspace{7pt}
\end{equation}
where $\mathcal{G}_{zk}$ is the set of variable nodes which are the arguments of the local function $f_k$ at the factor node $k$. This message transfer is illustrated in the right half of Fig.~\ref{fig:message_xchange}. Further, $R_{zk}^{(\nu-1)}$ is a value between zero and one and can be calculated as follows:
\begin{equation}
    R_{zk}^{(\nu-1)} = 1-\frac{1}{\rho|\mathbf{N_k}|}\sum\limits_{i\in\mathbf{N_k}}\sum\limits_{x\in\Upsilon}|T_{ki}-x|\mu^{(\nu-1)}_{i\rightarrow{k}}(x).
    \label{eqn:lambda_zero_2}
    \vspace{-3pt}
\end{equation}
The above equation can be interpreted as one minus the average inconsistency of user $k$ calculated by using the messages it received from all its neighbors. Further, $\rho$, which is the highest possible deviation of a user, is set to $4$ in this particular rating system, where the rating values are integers from the set $\Upsilon$. Thus, the reliability of users (in their ratings) is measured based on the messages formed by the algorithm. Using (\ref{eqn:lambda_zero}) and assuming that the predicted ratings in set $\mathcal{G}_{zk}$ are independent from each other at each intermediate step (to reduce the computational complexity), it can be shown that
\begin{align}
    f_k(\mathcal{G}_{zk}^{(\nu)},\mathbb{T}_k,R_{zk}^{(\nu-1)})=\prod\limits_{i\in\mathbf{N_k}}f_k(G_{zi}^{(\nu)},\mathbb{T}_k,R_{zk}^{(\nu-1)}).
    \label{eqn:intermediate1}
   \end{align}
Thus, the message in (\ref{eqn:lambda_zero}) becomes
\begin{align}
    &\lambda^{(\nu)}_{k\rightarrow{a}}(G_{za}^{(\nu)}) = \hspace{-0pt} f_k(G_{za}^{(\nu)},\mathbb{T}_k,R_{zk}^{(\nu-1)})\times \nonumber \\ &\Bigl\{\sum\limits_{\mathcal{G}_{zk}^{(\nu)}\backslash\{G_{za}^{(\nu)}\}}\hspace{-0pt}\Bigl[\prod\limits_{i\in\mathbf{N_k}\backslash\{a\}}f_k(G_{zi}^{(\nu)},\mathbb{T}_k,R_{zk}^{(\nu-1)})\prod\limits_{x\in\Delta}\mu^{(\nu-1)}_{x\rightarrow{k}}(G_{zx}^{(\nu)})\Bigr]\Bigr\}.
    \label{eqn:intermediate2}
\end{align}
Since the second part of (\ref{eqn:intermediate2}) is a constant, $\lambda^{(\nu)}_{k\rightarrow{a}}(G_{za}^{(\nu)}) \propto f_k(G_{za}^{(\nu)},\mathbb{T}_k,R_{zk}^{(\nu-1)})$, and hence, $\lambda^{(\nu)}_{k\rightarrow{a}}(G_{za}^{(\nu)}) \propto p(G_{za}^{(\nu)}|T_{ka},R_{zk}^{(\nu-1)})$, where\vspace{7pt}
\begin{align}
    &p(G_{za}^{(\nu)}=\ell|T_{ka},R_{zk}^{(\nu-1)}) = \nonumber \\ \nonumber \\
    &\begin{cases}
    \Bigl[R_{zk}^{(\nu-1)} + (1-R_{zk}^{(\nu-1)})\times\frac{|\kappa_a^z(\ell)|+1}{\sum\limits_{h\in\Upsilon}[|\kappa_a^z(h)|+1]}\Bigr] & \text{if $T_{ka}=\ell$} \\\\
    \Bigl[(1-R_{zk}^{(\nu-1)})\times\frac{|\kappa_a^z(T_{ka})|+1}{\sum\limits_{h\in\Upsilon}[|\kappa_a^z(h)|+1]}\Bigr] & \text{if $T_{ka}\neq\ell$.}
    \end{cases}
    \label{eqn:lambda_zero_1_RS}
\vspace{-.1in}
\end{align}
Here, $\kappa_a$ denotes the genre (i.e., type) or the set of genres of item $a$. Further, $|\kappa_a^z(h)|$ is the number of items in the same genre as $\kappa_a$ which are previously rated as $h$ by the active user $z$. The way we compute the probabilities in (\ref{eqn:lambda_zero_1_RS}) resembles the belief/plausibility concept of the Dempster-Shafer Theory~\cite{Shafer-1976}. Given $T_{ka} = 1$, $R_{zk}^{(\nu-1)}$ can be viewed as the belief of user $k$ that $G_{za}^{(\nu)}$ is one (at the $\nu^{th}$ iteration). In other words, in the eyes of user $k$, $G_{za}^{(\nu)}$ is equal to one with probability $R_{zk}^{(\nu-1)}$. Thus, $(1-R_{zk}^{(\nu-1)})$ corresponds to the uncertainty in the belief of user $k$. In order to remove this uncertainty and express $p(G_{za}^{(\nu)}|T_{ka},R_{zk}^{(\nu-1)})$ as the probabilities that $G_{za}^{(\nu)}$ is $\ell$ ($\ell\in\Upsilon$), we distribute the uncertainty among the possible outcomes (one to five) in proportion to the histogram of the ratings provided by the active user $z$ for the items in the same genre as $\kappa_a$. That is, if the active user previously provided high ratings for the items in the same genre as $\kappa_a$, then we distribute most of the uncertainty to the higher ratings in proportion to the rating histogram of the active user for the items in the same genre as $\kappa_a$. Similarly, if the active user previously provided low ratings for the items in the same genre as $\kappa_a$, we distribute most of the uncertainty to the lower ratings. Therefore, from user $k$'s point of view, $G_{za}$ is equal to one with probability $R_{zk}^{(\nu-1)} + (1-R_{zk}^{(\nu-1)})\times\frac{|\kappa_a^z(1)|+1}{\sum\limits_{h\in\Upsilon}[|\kappa_a^z(h)|+1]}$. On the other hand, it is equal to $\ell$ ($\ell\neq1$) with probability $(1-R_k^{(\nu-1)})\times\frac{|\kappa_a^z(\ell)|+1}{\sum\limits_{h\in\Upsilon}[|\kappa_a^z(h)|+1]}$. We note that the above discussion assumed $T_{ka}=1$ and similar statements hold for the cases when $T_{ka} = 2,3,4,5$. It is worth clarifying that, as opposed to the Dempster-Shafer Theory, we do not combine the beliefs of the users. Instead, we consider the belief of each user individually and calculate probabilities that $G_{za}^{(\nu)}$ being $\ell$ ($\ell\in\Upsilon$) in the eyes of each user as in (\ref{eqn:lambda_zero_1_RS}). We note that if the active user $z$ did not rate any items from this particular genre ($\kappa_a$), we distribute the uncertainty in proportion to the average rating of user $z$ (for the items it previously rated) ($A_z=\frac{\sum_{i\in\mathbf{N_z}}T_{zi}}{|\mathbf{N_z}|}$). The above computation in (\ref{eqn:lambda_zero_1_RS}) must be performed for every neighbors of each factor node. This finishes the first half of the $\nu^{th}$ iteration.
In the second half of the ${\nu}^{th}$ iteration, we calculate the message $\mu^{(\nu)}_{a\rightarrow{k}}(G_{za}^{(\nu)})$ by multiplying all probabilities the variable node $a$ received from its neighbors excluding the one from the factor node $k$, as shown in the left half of Fig.~\ref{fig:message_xchange}. We note that the previous ratings of the active user play a key role in the algorithm. Hence, the values of those variables in $\mathbb{G}_z$ which are associated with the items already rated by the active user $z$ are set to the corresponding ratings (i.e., $G_{zj}=T_{zj}$ if $j\in\mathbf{N_z}$). Thus, if $a\in\mathbf{N_z}$, the messages generated from the variable node $a$ do not vary with iterations since the value of this variable node ($G_{za}$) is fixed based on the ratings of the active user. Therefore, the message from the variable node $a$ to the factor node $k$ at the ${\nu}^{th}$ iteration is given by
\begin{align}
    &\mu^{(\nu)}_{a\rightarrow{k}}(G_{za}^{(\nu)}=\ell) = \nonumber \\ \nonumber \\
    &\begin{cases}
\vspace{-.2in}
    \frac{1}{\sum\limits_{h\in\Upsilon}\prod\limits_{i\in\Xi}\lambda^{(\nu)}_{i\rightarrow{a}}(h)} \times \prod\limits_{i\in\Xi}\lambda^{(\nu)}_{i\rightarrow{a}}(G_{za}^{(\nu)}) & \text{if $a\not\in\mathbf{N_z}$} \\\\
    1 & \text{if $a\in\mathbf{N_z}$ and $T_{za}=\ell$} \\
    0 & \text{if $a\in\mathbf{N_z}$ and $T_{za}\neq \ell$.}
    \end{cases}
    \label{eqn:mu_zero_one_RS}
    \vspace{7pt}
\end{align}

The algorithm proceeds to the next iteration in the same way as the ${\nu}^{th}$ iteration. We clarify that the iterative algorithm starts by computing $\lambda^{(1)}_{k\rightarrow{a}}$ by using $R_{zk}^{(0)}=\varrho$, where $\varrho$ ($0<\varrho<1$) is the system's present confidence on the users for the reliability of their ratings computed at the previous execution of the algorithm. At the end of each iteration, the upper equation in (\ref{eqn:mu_zero_one_RS}), after following modification, is used to compute the prediction of ratings of the active user $z$. That is, we use the set $\mathbf{N_a}$ instead of $\Xi$ in (\ref{eqn:mu_zero_one_RS}) to compute $\mu_a^{(\nu)}(G_{za}^{(\nu)})$ for every item $a$ for which the active user $z$ did not have any rating. Then, we set $G_{za}^{(\nu)} = \sum_{i=1}^5{i\mu_a^{(\nu)}(i)}$. The iterations stop when $G_{zj}$ values converge for every item $j$.

\vspace{-.05in}
\section{Evaluation of BPRS}\label{sec:evaluation}

We evaluate the performance of BPRS using the $100$K MovieLens dataset. The dataset contains $100,000$ ratings from $943$ users on $1682$ items (movies) in which each user has rated at least $20$ items. Further, the rating values are integers from $1$ to $5$.  We note that based on our simulations, we observed that BPRS converges, on the average, in $10$ iterations. Therefore, for the remaining of this section, we either show our results during the first $10$ iterations or after the $10^{th}$ iteration.
\vspace{-.1in}
\subsection{Prediction Accuracy}

We evaluate the rating prediction accuracy of BPRS in terms of Root Mean Square Error (RMSE) metrics over the predicted ratings. We note that each test dataset is created by $80\%/20\%$ split of the full data into training and test data.Then, we used the training data ($80\%$ of the whole dataset) to predict the ratings in the test dataset. We computed the RMSE  as below:

\vspace{-.1in}
\begin{equation}
    \mathrm{RMSE}=\sqrt{\frac{1}{|K|}\sum\limits_{i\in\mathbb{U},j\in\mathbb{I}}(G_{ij}-\hat{G}_{ij})^2}
    \label{eqn:RMSE_dfn}
\end{equation}
where $|K|$ is the number of ratings (to be predicted) in the test dataset, $\hat{G}_{ij}$ is the actual value of the rating provided by user $i$ for the item $j$ in the test dataset, and $G_{ij}$ is the predicted rating value by the algorithm.

\begin{figure}[t]
   \centering
   \small
   \hspace{-10pt}
   \vspace{-14pt}
  \epsfig{width=.9\columnwidth,file=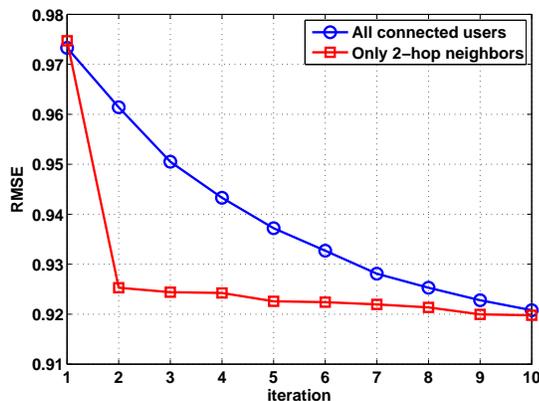}
      \caption{Performance of BPRS in RMSE vs. number of iterations when: (i) all users and (ii) only the 2-hop neighbors are used.}
      \label{fig:sampling_RMSE_recsys}
\vspace{-.2in}
\end{figure}

In Figs.~\ref{fig:sampling_RMSE_recsys}, we show the RMSE  provided by BPRS for two different scenarios: when all users connected to each active user via a path are used and when only the 2-hop neighbors of each active user are used in the algorithm. We observe that keeping only the 2-hop neighbors of each active user provides better performance in terms of RMSE.

%

Finally, we evaluated BPRS against some popular recommendation algorithms such as: 1. MovieAvg (which computes the predicting ratings for the movies by averaging all the received ratings for each movie) with an RMSE of $1.053$, 2. Correlation-based neighborhood model (CorNgbr), with an RMSE of $0.9406$~\cite{Koren_2008}, and 3. SVD latent factor model, with $50$ factors and RMSE of $0.9046$~\cite{Koren_2008}. We conclude that BPRS is comparable to existing methods such as CorNgbr and SVD in terms of rating prediction accuracy. On the other hand, BPRS generates recommendations in linear complexity for each active user and updates the recommendations for each active user instantaneously using the most recent data.

\subsection{Computational Complexity}

Assuming $u$ users and $s$ items in the system, we obtained the computational complexity of BPRS (in the number of multiplications) as $\mathbf{max}(\mathbb{O}(cs),\mathbb{O}(cu))$ per each active user, where $c$ is the average number of nonzero elements in each row of the user-item matrix. We note that due to the sparseness of the user-item matrix, the coefficient $c$ is a small number. Further, as we discussed before, BPRS converges, on the average, in $10$ iterations. Hence, we did not include the number of iterations in the complexity measure as it only introduces a small constant in front of the total complexity. This result indicates that BPRS can compute the recommendations for each active user very efficiently using the most recent data (ratings). Therefore, we claim that the BP-based approach toward the recommendation problem is very promising and can result in a new class of accurate and scalable recommender systems.

\vspace{-.1in}
\section{Conclusion}
\label{sec:conclusion}

In this paper, we introduced the Belief Propagation Based Iterative Recommender System (BPRS). BPRS formulates the recommendation problem as making statistical inference about the ratings of users for unseen items based on observations. BPRS provides a complexity that remains linear per single active user, making it very attractive for large-scale systems. Further, it can update the recommendations for each active user instantaneously using the most recent data (ratings) and without solving the recommendation problem for all users. While providing these significant scalability advantages over the existing methods, we showed that BPRS also provides comparable rating prediction accuracy with popular methods.

\vspace{-.1in}

\bibliographystyle{IEEEtran}
\bibliography{reputation,current_proposal,past_nsf_grant_publication,recommender_systems}

\end{document}